%% file: main.tex
\documentclass[runningheads]{llncs}
\usepackage{graphicx}
\usepackage{subcaption} 
\usepackage{caption}
\usepackage{comment}

\usepackage{booktabs}
\usepackage{multirow}
\usepackage{makecell}

\usepackage{amsmath,amssymb} 
\usepackage{color}
\usepackage{hhline,arydshln}
\usepackage[pagebackref=true,breaklinks=true,colorlinks,bookmarks=false]{hyperref}
\usepackage{orcidlink}

\usepackage[accsupp]{axessibility}

\begin{document}
\pagestyle{headings}
\mainmatter
\def\ECCVSubNumber{29}  

\title{DSR: Towards Drone Image Super-Resolution} 

\titlerunning{Towards Drone Image Super-Resolution}

\author{Xiaoyu Lin\inst{}\orcidlink{0000-0003-3240-3683} \and
Baran Ozaydin\inst{} \and
Vidit Vidit\inst{} \and
Majed El Helou\inst{}\orcidlink{0000-0002-7469-2404} \and
Sabine S\"usstrunk\inst{}\orcidlink{0000-0002-0441-6068}
}

\authorrunning{X. Lin et al.}

\institute{School of Computer and Communication Sciences, EPFL, Switzerland
\email{\{xiaoyu.lin,baran.ozaydin,vidit.vidit,majed.elhelou,sabine.susstrunk\}@epfl.ch}
}
\maketitle

\begin{abstract}
Despite achieving remarkable progress in recent years, single-image super-resolution methods are developed with several limitations. Specifically, they are trained on fixed content domains with certain degradations (whether synthetic or real). The priors they learn are prone to overfitting the training configuration. Therefore, the generalization to novel domains such as drone top view data, and across altitudes, is currently unknown. Nonetheless, pairing drones with proper image super-resolution is of great value. It would enable drones to fly higher covering larger fields of view, while maintaining a high image quality. 

To answer these questions and pave the way towards drone image super-resolution, we explore this application with particular focus on the single-image case. We propose a novel drone image dataset, with scenes captured at low and high resolutions, and across a span of altitudes. Our results show that off-the-shelf state-of-the-art networks witness a significant drop in performance on this different domain. We additionally show that simple fine-tuning, and incorporating altitude awareness into the network's architecture, both improve the reconstruction performance. 
\footnote{Our code and data are available at \url{https://github.com/IVRL/DSR}.}

\end{abstract}

\section{Introduction}

\begin{figure}[t]
    \centering
    \includegraphics[width=0.98\linewidth]{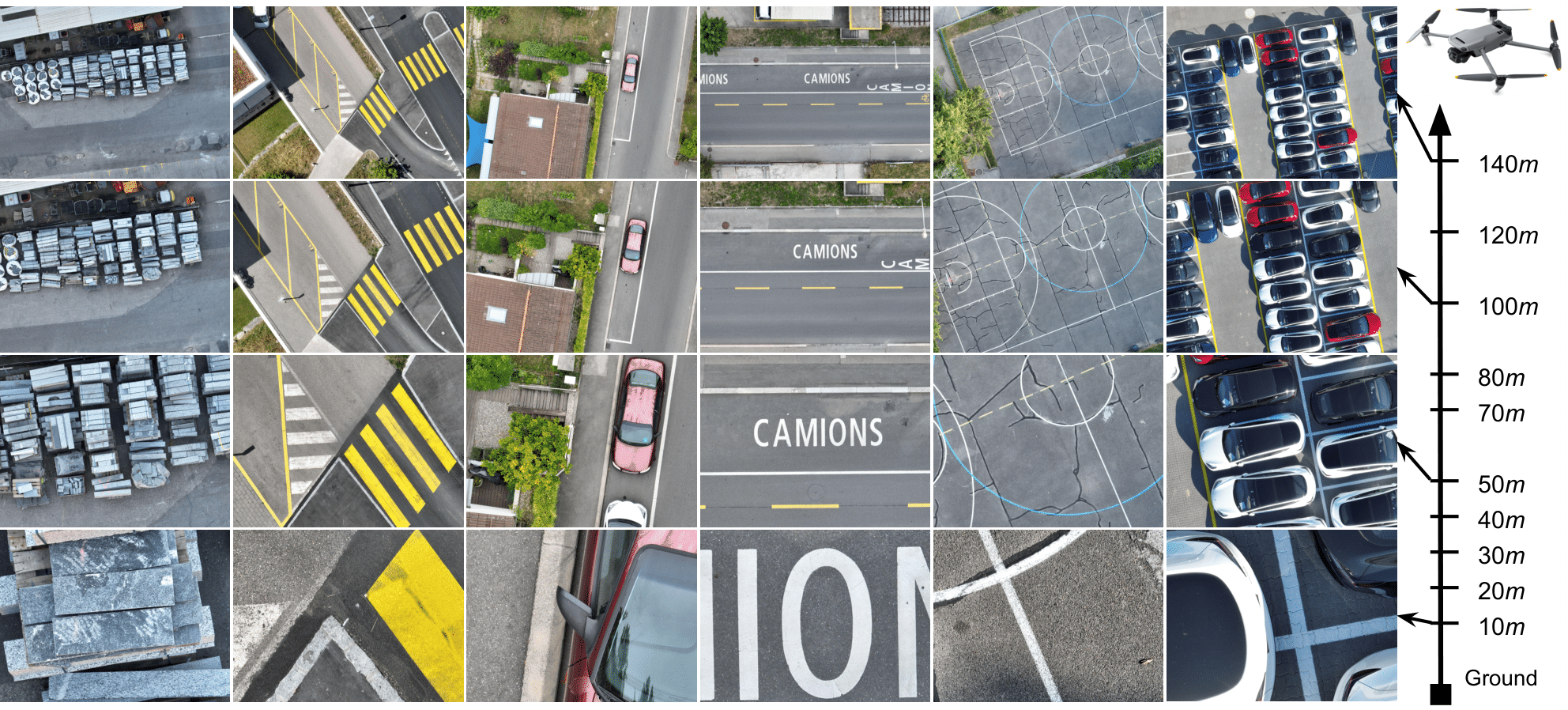}
    \caption{Sample images from our DSR dataset. The drone takes off at ground level, moves to a predetermined set of altitudes, and captures images at each of these altitudes. 
    }
    \label{fig:my_label}
\end{figure}

The task of single-image super-resolution (SR) is the reconstruction of a high-resolution (HR) image from a single low-resolution (LR) capture. Super-resolution is widely studied in the literature and has witnessed significant progress with deep learning approaches. Multiple HR images can correspond to a single LR observation, making the SR task an ill-posed problem. In other words, SR methods are bound to learn an image distribution prior to reconstruct an \textit{expected} HR image. With the power of deep networks to implicitly learn prior distributions, SR performance improved and it opened up new possibilities to push the limits of imaging systems.

Although deep-learning-based SR methods pushed forward the state-of-the-art performance, multiple constraints were incorporated along the way. The models are trained with certain assumptions on the degradation model that are learned during training. Furthermore, the underlying datasets are human-captured and thus suffer from a content bias. Namely, these datasets are typically captured along the horizontal direction and at a human-level height. Thus, the methods trained on this domain do not generalize well to different data such as top view imaging. 

Top view images constitute, however, most of what a drone would observe in the majority of drone applications. Drones can significantly benefit from SR solutions as they would enable the drone to fly at higher altitudes, survey wider areas, and preserve good image quality. However, because of the necessary reliance on data priors, it is unclear whether available SR solutions would perform well on drone images, and whether a domain gap exists across altitudes. 

In an effort towards enabling drone image super-resolution, we create the \textbf{Drone Super-Resolution}, \textbf{DSR}, dataset. It is the first dataset of drone images containing low-resolution and high-resolution drone-based image pairs, with varying altitudes. We obtain our high-resolution targets with optical zoom, and we collect 10 versions of each scene for 10 different chosen altitudes, \ref{fig:my_label}. With our DSR benchmark, we evaluate the performance of available SR methods. We observe lower performance on our drone data, and a difference in performance between altitudes. We show in the frequency domain that a domain gap also exists for different altitudes of drone capture. We finally propose a solution that exploits altitude meta data. We show that this altitude-aware approach, similarly to fine-tuning, outperforms previous SR methods, even degradation-adaptive ones.
Our contributions can be summarized as follows:
\begin{itemize}
    \item We introduce a novel dataset, DSR, for drone image super-resolution. Our dataset provides pairs of low-resolution and high-resolution images. Furthermore, we capture each scene at 10 different altitudes. We also provide a burst of 7 low-resolution images and RAW data to encourage future research. 
    \item We analyze domain variability, its effect on regular data vs. top view data, and also across altitudes for the drone data. We benchmark nine off-the-shelf super-resolution methods on DSR.
    \item We show that an altitude-aware architecture, as well as fine-tuning, can outperform previous methods.
\end{itemize}

\section{Related work}
\textbf{Drone imaging.} 
Various novel applications rely on drones and drone imaging. The research can be divided into two categories. The first category comprises the methods designed for the drones themselves (e.g. for localization or autonomous navigation), and the second are methods developed with drone data for downstream tasks (e.g. surveillance or traffic control).

Computer vision methods already play an essential role already within drones, for instance, camera calibration and image matching~\cite{xu2016mosaicking} for drones' automated flight and stabilization, including pose estimation and obstacle detection~\cite{al2018survey}. 
Many more industrial applications exist that exploit images and videos captured by drones. Some drone-based image or video datasets were developed for specific tasks such as traffic and crowd detection~\cite{oh2011large}, notably for surveillance~\cite{liu2015fast}. 
Other datasets were proposed for the object detection task~\cite{hsieh2017drone,xu2019dac}. The UAVDT~\cite{du2018unmanned} benchmark forms a database for object detection and tracking using drone images. UAVDT also contains some real-world challenges, such as varying weather conditions, flying altitudes, and drone views. 
These datasets are used to evaluate available approaches and develop novel methods, especially for learning-based solutions. However, the publicly available datasets focus on object detection and other related computer vision tasks. Here, we present DSR, which is a drone dataset dedicated to the super-resolution task. Rather than sampling video frames~\cite{du2018unmanned}, we opt for the more time-consuming approach of capturing one image at a time in order to obtain the highest quality possible. DSR consists of paired low-resolution and high-resolution images, across varying altitudes, and includes burst sequences and RAW data. 

\textbf{Super-resolution.}
SR methods are based on edge model assumptions~\cite{chan2009neighbor}, priors on gradient distributions~\cite{sun2008image}, or learn from examples as with deep neural networks which hold the state-of-the-art reconstruction performance. These networks exploit deep architectures with residual learning and attention mechanisms~\cite{zhang2018image}, or train over the frequency domain using wavelet transforms for memory and compute optimization~\cite{zhou2019comparative}. A variety of other solutions were developed to exploit perceptual losses~\cite{johnson2016perceptual} or GAN models~\cite{ledig2017photo,wang2018esrgan} to improve the qualitative performance of the SR networks. However, the SR networks face generalization issues as highlighted in~\cite{shocher2018zero}, notably in their learned priors~\cite{el2020stochastic}. Despite attempts to account for degradation changes~\cite{SRMD}, the generalization still suffers because of the need to estimate blur kernels (an estimation which is, itself, prone to over-fitting).
Recently, two datasets were built using optical zoom with varying focal lengths to obtain paired low-resolution and high-resolution image pairs~\cite{cai2019toward,zhang2019zoom}. These datasets all include similar content from a unique domain. This is due to the fact that their images were captured from a human-perspective, with generally uniform height and a camera axis along the horizontal direction. We study with DSR the application of SR to a different domain with a different image prior, for drone images captured from a top view and at varying altitudes. We study the performance of available methods on this novel domain, and propose solutions to improve their results. We hope DSR will foster future research towards developing drone image super-resolution.

\section{DSR dataset}
We create our DSR dataset by capturing image pairs of the same scene with different focal length values. We take photos at ten different altitudes for each scene. Our altitude values form the set \{10, 20, 30, 40, 50, 70, 80, 100, 120, 140\}$m$, which is chosen in such a way as to have good sampling across altitudes, various altitudes that are multiples of each other, and to span a large range of altitudes that are relevant for drone applications~\cite{du2018unmanned}. The dataset is split into three sets for training, validation, and testing, while ensuring that no overlaps exist between any of the sets (Table~\ref{tab: statistic}). We choose a large size of validation and test sets relative to the training set (one to two ratio), to emphasize the quality of the empirical evaluation. We also show in Table~\ref{tab: statistic} that the final number of low-resolution and high-resolution pairs is evenly distributed across altitudes and does not cause any data imbalance. 


\input{tables/statistics}

\subsection{Data acquisition}
\textbf{Localization.} We use a DJI Mavic 3 drone for collecting the images of our DSR dataset. For controlling the altitude, the vertical hovering accuracy range of the drone when using GNSS (GPS+Galileo+BeiDou) is $\pm0.5m$. We exploit this information as an altitude guide and regularly refine our estimates with a measurement from the onboard barometer, which is measuring atmospheric pressure change relative to the take-off ground position and can provide improved accuracy. We additionally rely on GNSS for horizontal positioning to ensure no overlap between acquired images, based on the drone's geographic location. 

\textbf{Cameras.} The drone comes pre-equipped with two cameras that are very closely mounted. The first is a 4/3-in CMOS sensor Hasselblad L2D-20c camera (referred to hereinafter as "Hasselblad camera"). The second is a 1/2-in CMOS sensor Tele camera (referred to hereinafter as "Tele camera"). The focal length of the Hasselblad camera is $24mm$, and that of the Tele camera is $162mm$. 
The Hasselblad camera captures images of $5280\times3956$ pixel resolution, and the Tele camera captures images of $4000\times3000$ pixel resolution~\cite{drone}. The Tele camera has a larger focal length, meaning the images captured by the Tele camera form an optically zoomed version of the ones captured with the Hasselblad camera at the same position. 
Therefore, images from the Tele camera are used to generate the high-resolution ground-truth target data and matched images from the Hasselblad camera are used to generate their low-resolution counterpart.

\textbf{Capture settings.} The drone's Tele camera has a fixed f-number, optimized and set by the manufacturer to 4.4. This aperture size yields a good trade-off between depth of field and exposure. A small aperture size is advantageous for super-resolution data acquisition~\cite{zhang2019zoom,cai2019toward}, as it extends the depth of field and hence the range of depth values in the captured scene that are only minimally blurred. However, exposure is also critical for a drone because of the inherent instability in such applications. As the drone is exposed to both its own vibrations as well as to the wind, a large enough aperture size enables a reduction in exposure time and hence a reduction in motion blur. 

\textbf{Acquisition details.} We match the exposure value between the Tele camera and the Hasselblad camera in real-time during capture onboard the drone. The Hasselblad camera settings (f-number and shutter speed) are adjusted on the drone to match the exposure value of the Tele camera capture.
We thus begin by shooting with the Tele camera, then the Hasselblad camera. With the latter, we shoot a burst of 7 consecutive images. We include this burst capture for future research, as multi-frame super-resolution has the potential to achieve better results than single image super-resolution~\cite{bhat2021deep,lecouat2021lucas,dudhane2021burst} despite its practical inconvenience. 
For each scene, i.e. horizontal geographic location, we capture at each of \{10, 20, 30, 40, 50, 70, 80, 100, 120, 140\}$m$ altitudes a Tele camera image and 7 Hasselblad images, while preserving the exposure value between paired captures. With this configuration, we follow the convention in BurstSR~\cite{bhat2021deep} and capture 200 scenes in total, which we split into non-overlapping train, validation, and test sets consisting of 100, 50, and 50 scenes, respectively. Every \textit{scene} is the equivalent of 8 captures $\times$ 10 altitudes. After creating 180$\times$180 resolution patches~\cite{bhat2021deep}, this yields a total of 5175, 2622, and 2532 image pairs for training, validation, and test sets. Our patches are extracted from the LR images in a sliding window manner. To avoid overlap among patches, we extract 180$\times$180 patches from LR images with a stride of 180 pixels.



\subsection{Image registration}

For RAW data, we first pack each 2 × 2 block in the raw Bayer color filter array along channel dimensions, to obtain 4-channel images. In all following data processing steps, we obtain the parameters of homography transformations based on the RGB image and apply the same transforms to both RGB and RAW images for consistency. 
For burst sequences, we obtain the parameters of transforms based on the first burst frame. It is worth noting that although we focus in this paper on RGB data and single-image (first frame) SR for their practicality in real-world applications, we also provide burst sequences and RAW data in DSR to advance future research on drone SR.

Since the images captured by two cameras have different fields of view (FOV), we first crop the matched FOV from each LR image in the burst sequence. In data acquisition, we use the two adjacent cameras to capture our HR and LR images. The vibration of a flying drone causes more severe displacements than, for instance, human held capture or tripods. Thus, simply cropping around the center of the LR image~\cite{zhang2019zoom,cai2019toward} cannot always achieve FOV matching. Similar to BurstSR~\cite{bhat2021deep}, we estimate the homography between the LR and HR paired images using SIFT~\cite{lowe1999object} and RANSAC~\cite{fischler1981random} to perform FOV matching. However, in our dataset, the can drone can move subtly not only horizontally but also vertically, which leads to small variations in the corresponding FOV on the LR image. Although the worst variations are only a few pixels wide, they would complicate downstream data possessing steps by causing a varying scale factor, albeit by an almost negligible magnitude. To overcome this problem, we resize the FOV to a fixed resolution (720$\times$540). Therefore, the scale factor of our DSR dataset is always constant at exactly $\times50/9$. We choose the nearest-neighbor interpolation method to resize the matched FOV. This choice is based on the fact that the majority of pixel values are not even affected because the worst dimension variations are only a few pixels wide within the 720$\times$540 resolution, and most images do not have any variation. A detailed comparison including other resizing methods is provided in the supplementary material. 


\textbf{Exposure.} Despite matching the same exposure value across captures, we still found some color difference in certain image pairs. To improve the accuracy of our image pair registration, we exploit the pixel-wise color transfer algorithm of RealSR~\cite{cai2019toward}. Additionally, we apply histogram matching and evaluate both methods on our dataset (see supplementary material for further evaluation results). We thus apply color transfer and histogram matching for color correction on our paired images, to minimize any potential variations.

\begin{figure}[tb]
\centering
\subfloat[Examples from the DSR dataset at \\10$m$ and 50$m$, and the RealSR dataset.]
{\includegraphics[width=.495\columnwidth]{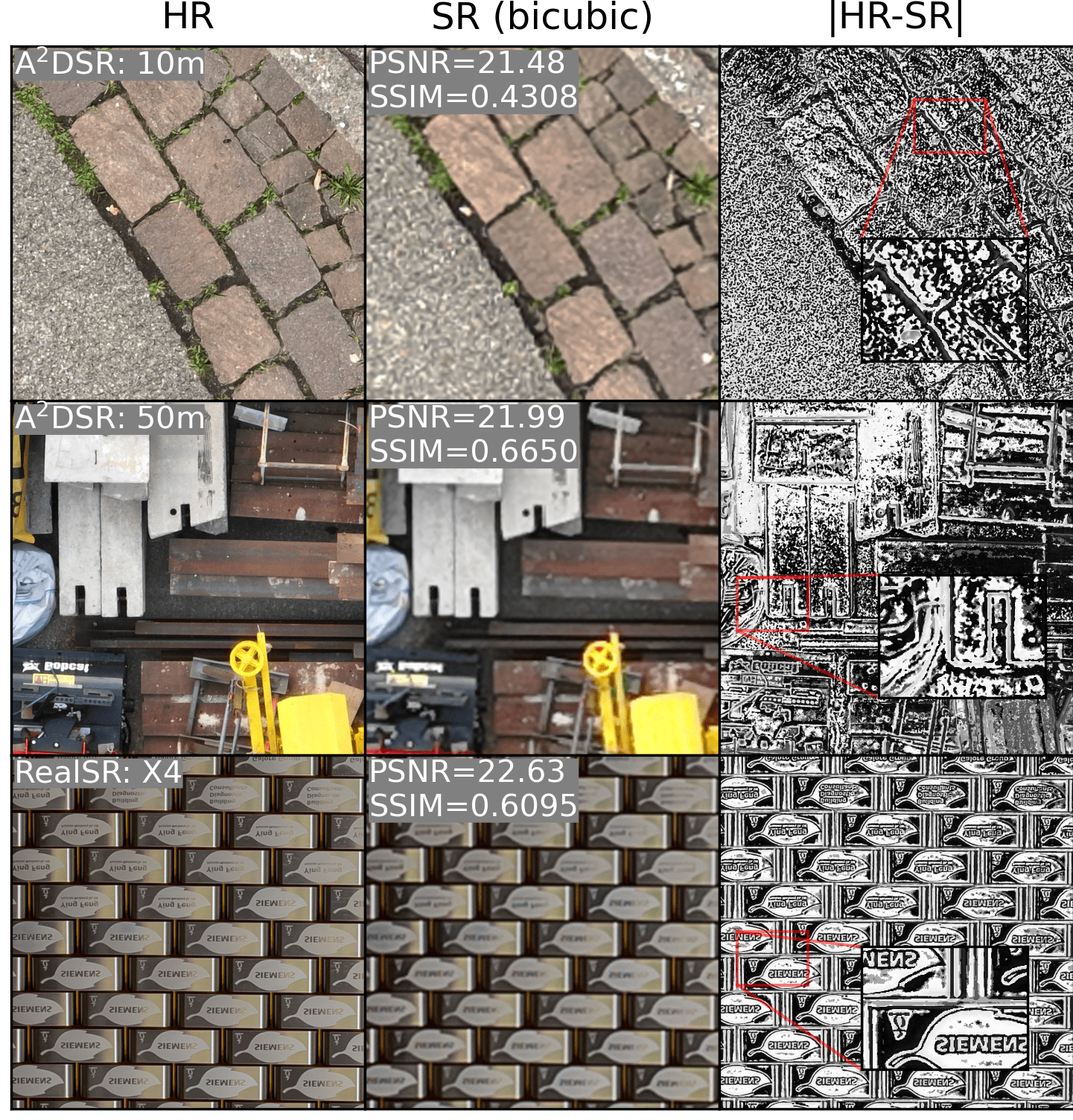}}
\subfloat[Examples from the DSR dataset at 70$m$ and 120$m$, and the RealSR dataset.]
{\includegraphics[width=.495\columnwidth]{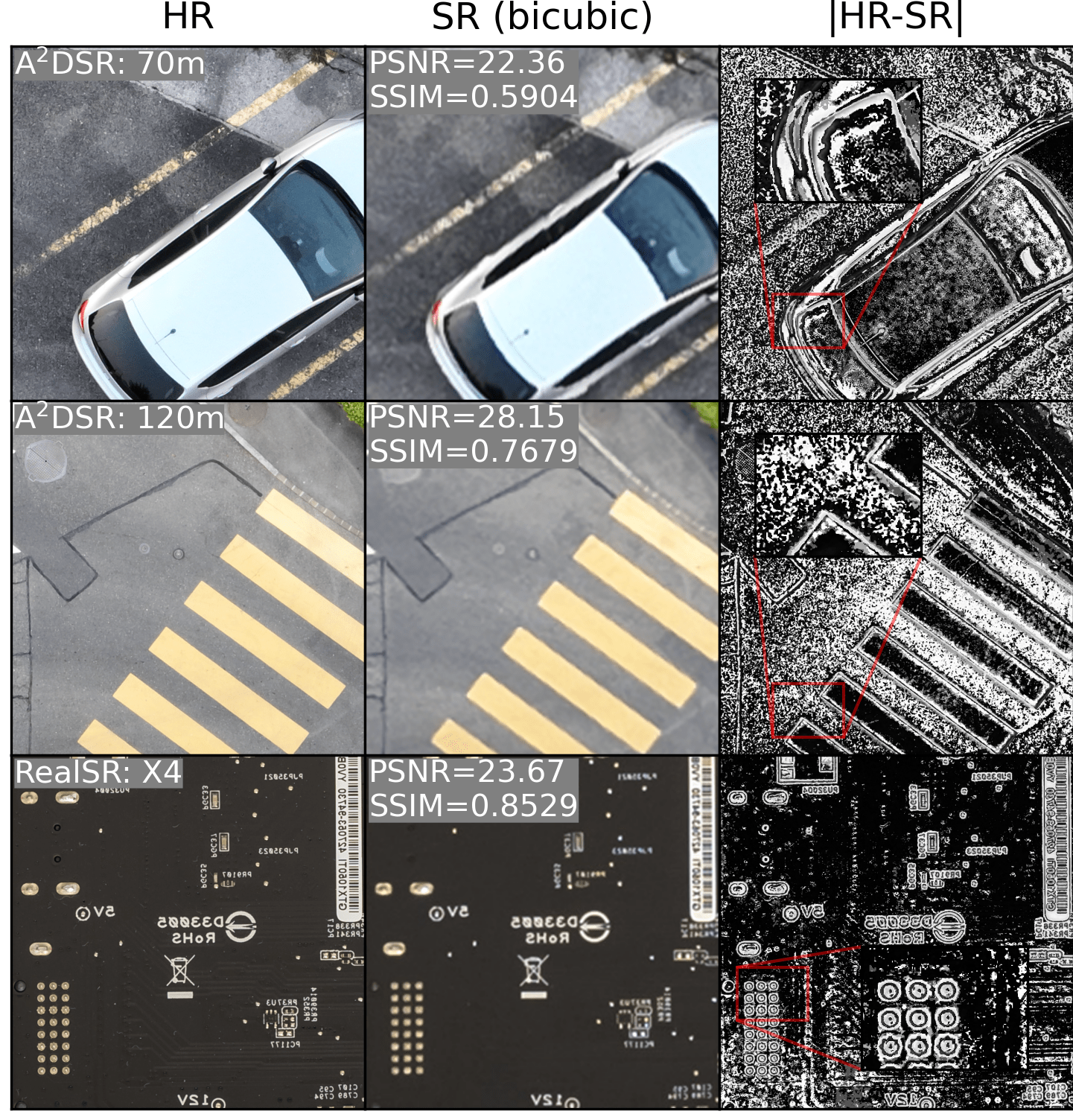}}
\caption{We analyze the registration on the higher resolution for more precision. The HR image is compared with the bicubic upsampled LR. We note that the absolute error between the two is around edges, which are sharper in the HR image. We can see in the second row of (b), that the errors on symmetric surfaces (yellow rectangles) are similarly symmetric and follow the same contour shape. This illustrates the alignment quality, as any misalignment would cause shape distortions between the image and error shape contours. For reference, the bottom row (a,b) shows a similar analysis on RealSR images~\cite{cai2019toward}.}
\label{fig: alignment}
\end{figure}

\begin{figure}[tb]
\centering
\subfloat[Frequency content distribution (HR).]
{\label{fig: psd}
\includegraphics[height=.335\columnwidth]{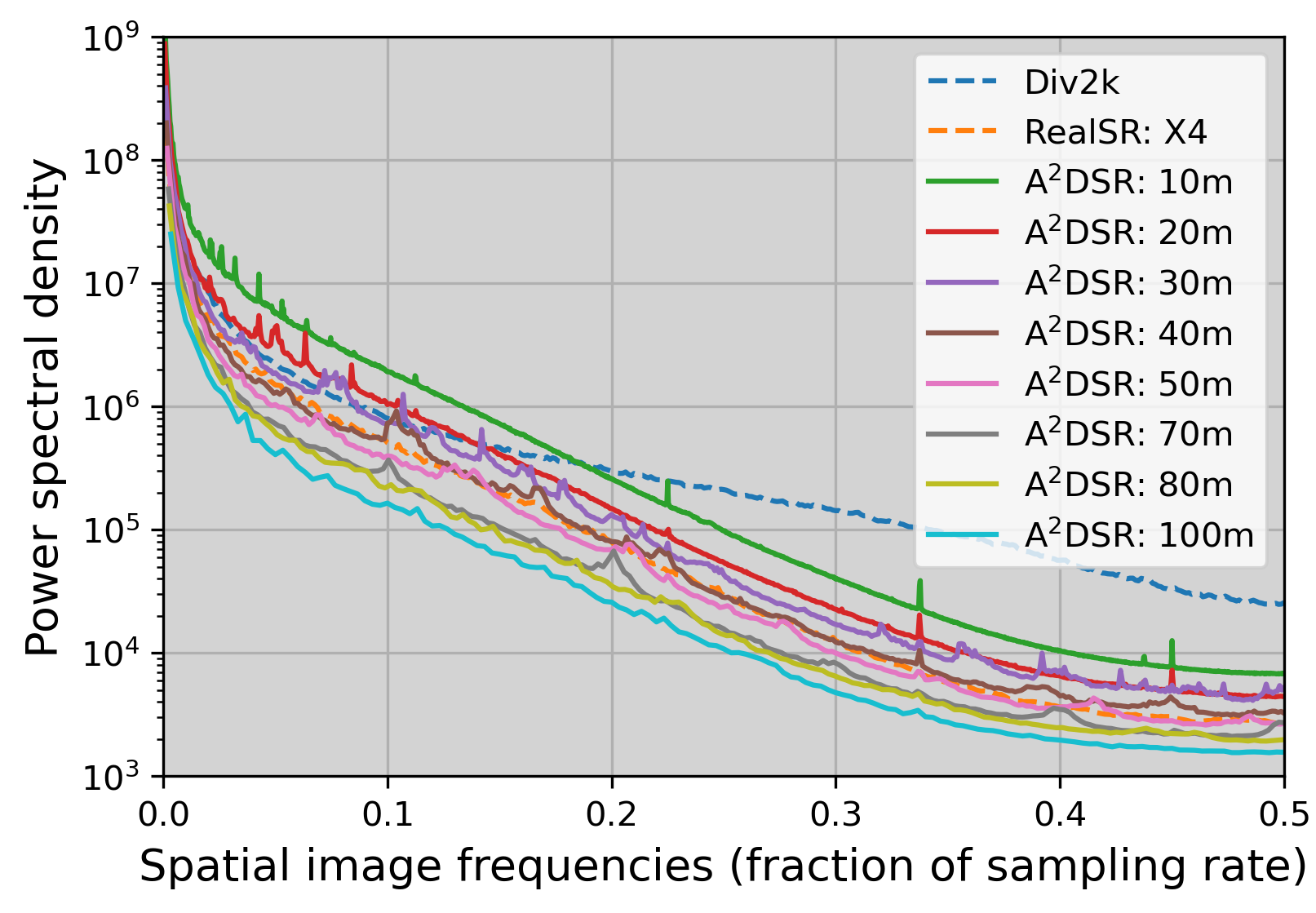}} \hfill
\subfloat[Estimated blur kernels.]
{\label{fig: kernel}
\includegraphics[height=.335\columnwidth]{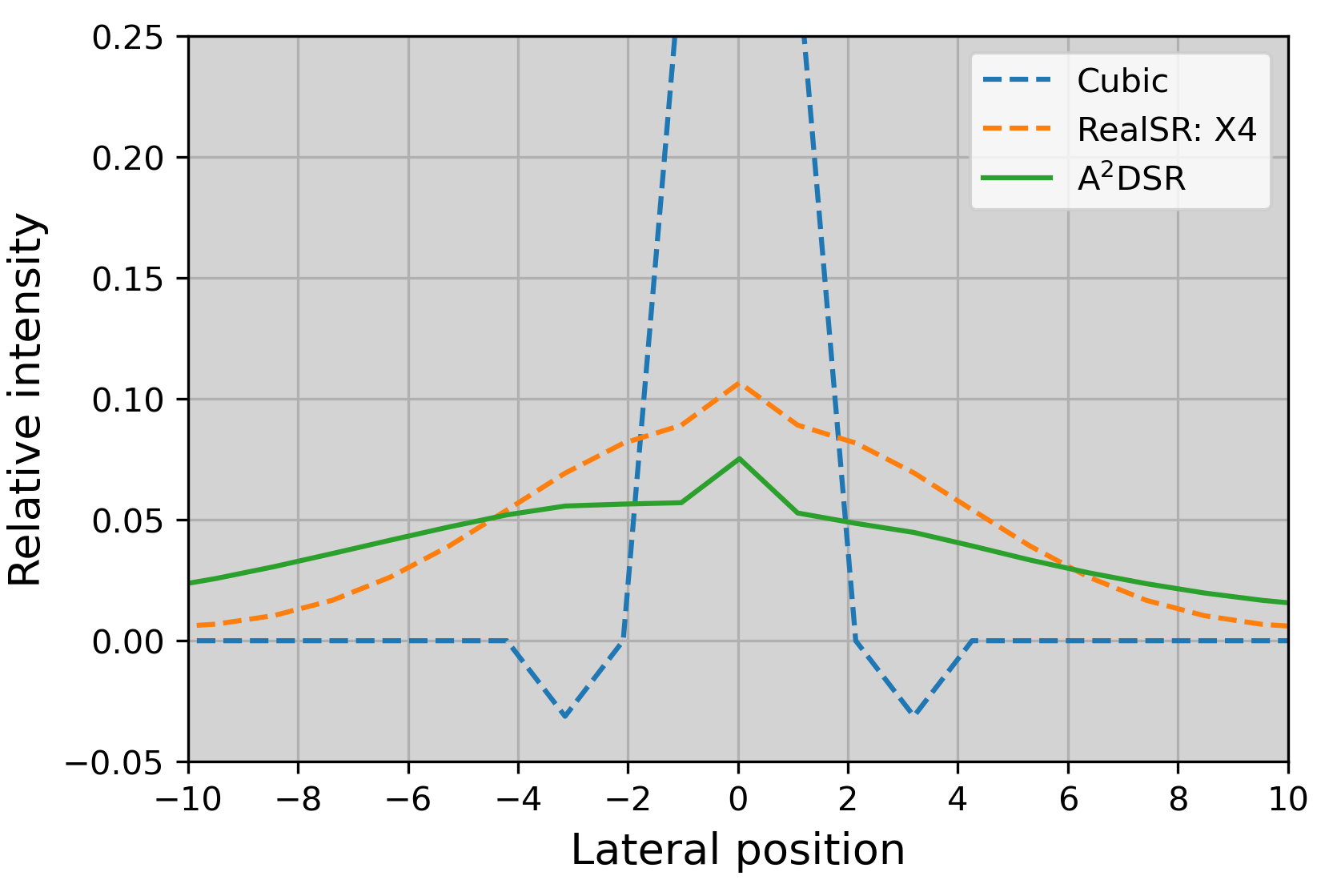}}
\caption{(a) Average PSD in different datasets and altitudes, and (b) blur kernel estimation. DSR has similar spatial frequency characteristics as other real-world datasets. The curve of lower altitude lies above the curve of higher altitude, the corresponding images contains more details. A longer tail in the blur kernel of DSR illustrates severe blur and thus harder-to-perform super-resolution.}
\label{fig: dataset analysis}
\end{figure}

\subsection{Alignment analysis}
A certain degree of misalignment cannot be avoided in super-resolution data acquisition, especially when working with drones. To best counter this problem, we carefully design external and internal control procedures.

\textbf{External control procedure.} We improve our alignment as well as possible by controlling external conditions. We acquire data in mild weather conditions with minimal wind and clouds, to reduce wind-induced movement and avoid moving cloud shadows in the scene. We carefully stabilize the drone at each altitude before capturing data. Furthermore, we look for static scenes and avoid moving vehicles or mobile objects. And lastly, we reduce the occurrences of vegetation especially trees with light leaves that can be moved by the wind.

\textbf{Internal control procedure.} For each patch, we reiterate the homography estimation between the LR patch and the corresponding region in the HR image to perform more refined local alignment. We align the LR image within the HR counterpart (rather than the reverse) for more precise alignment~\cite{bhat2021deep}. As a further control step, we remove all HR and LR pairs that have a normalized cross correlation lower than 0.9. The final numbers of \textit{valid} image pairs for each set at each altitude are listed in Table~\ref{tab: statistic}, and are balanced across all altitudes.


We visualize the absolute error map between the HR image and a bicubic upsampling of its LR image counterpart. The resulting mean spectral error value is visualized in Fig.~\ref{fig: alignment}. We can note that the absolute error over symmetric surfaces is consistently symmetric following the same contour shape, indicating a good alignment. The figure also shows in the last row a similar example from the RealSR~\cite{cai2019toward} that contains similar errors despite the use of a tripod during acquisition. These errors are due to the fact that LR images do not contain the sharp details of their HR counterparts, resulting in differences along edges~\cite{zhang2019zoom}.


\subsection{Data analysis}
Our DSR dataset contains image pairs captured at different altitudes. We analyze in this section, in the frequency domain, the potential domain gap across altitudes. 
We plot in Fig.~\ref{fig: psd} the average power spectral density (PSD) of our test set HR images for a set of altitudes and for two other datasets; the Div2K dataset~\cite{agustsson2017ntire} and a real-world SR dataset RealSR~\cite{cai2019ntire,cai2019toward}. To enforce FOV invariance across altitudes for our DSR dataset, we crop the center from higher altitudes. 
The PSD curves of DSR are closer to the curve of RealSR~\cite{cai2019ntire,cai2019toward}, while that of the Div2K dataset~\cite{agustsson2017ntire} is more flat with a slower drop along high frequencies. Div2K indeed contains high definition images with more high frequencies. 
What is most interesting to observe in the variation in our DSR dataset. We can see that the PSD at lower altitudes lies above the PSD of higher altitude images. This clearly highlights the existence of a domain gap, in terms of frequency content, across altitudes. As SR networks learn to predict conditional frequency bands~\cite{el2020stochastic}, it is important to note any differences in that domain. 

We also estimate the blur kernel between our LR and HR observations. We use the blur kernel estimation proposed in~\cite{zhou2020w2s}. We estimate the blur kernel on the test set of DSR, and on the RealSR dataset~\cite{cai2019ntire,cai2019toward} with scale factor of 4 for reference. We also plot the widely used MATLAB bicubic kernel. The results are visualized in Fig.~\ref{fig: kernel}. From this visualization, it is evident that the bicubic kernel significantly differs from the blur kernels of real-world datasets, as noted in~\cite{zhou2020w2s}. The blur kernel estimated from DSR is similar to the in RealSR but is even more different than the bicubic kernel. Its wider shape indicates stronger blur, and a more challenging SR reconstruction task on our DSR dataset.

\section{Evaluating SR on drone data}
We benchmark state-of-the-art methods on DSR. We evaluate the performance on each altitude to explore its effect. To look for potential domain gaps, and to give reference results, we also benchmark on commonly used public datasets. And, for consistent comparisons, all datasets are fixed to the same scale factor. We then evaluate fine-tuning and our proposed altitude-aware solutions.

\subsection{Pretrained off-the-shelf SR methods}
We use bicubic interpolation and eight state-of-the-art SR networks for benchmarking results: EDSR~\cite{lim2017enhanced}, RDN~\cite{zhang2018residual}, RCAN~\cite{zhang2018image}, ESRGAN~\cite{wang2018esrgan}, SwinIR~\cite{liang2021swinir}, BSRNet~\cite{zhang2021designing}, NLSN~\cite{Mei_2021_CVPR} and DASR~\cite{wang2021unsupervised}. We exclusively use the trained networks provided by the corresponding authors with the scale factor of 4. Out of these SR networks, ESRGAN, SwinIR and DASR were trained on both the Div2K~\cite{agustsson2017ntire} and Flickr2K~\cite{lim2017enhanced,timofte2017ntire} datasets.
BSRNet was trained on the Div2K~\cite{agustsson2017ntire}, Flickr2K~\cite{lim2017enhanced,timofte2017ntire}, WED~\cite{ma2016waterloo}, and FFHQ~\cite{karras2019style} datasets. The rest of the networks were trained on  the Div2K~\cite{agustsson2017ntire} dataset. To address the real-world scenario, SwinIR, NLSN and DASR exploit a designed degradation model applied to HR images when generating the LR counterparts. The rest of the networks only apply the standard degradation model (i.e., the MATLAB bicubic downsampling).

\input{tables/benchmark_public}

\input{tables/benchmark_our}

\input{tables/results}

We first benchmark on five commonly used public datasets: Div2K~\cite{agustsson2017ntire}, Set5~\cite{bevilacqua2012low}, Set14~\cite{zeyde2010single}, B100~\cite{martin2001database}, and Urban100~\cite{huang2015single}. The LR images are generated with MATLAB's bicubic downsampling with DSR's scale factor (i.e. $\times50/9$). Since the SR networks have a scale factor of 4, we upscale the outputs of the networks to the target size using bicubic interpolation for consistent evaluation. Following the method originally presented in~\cite{zhang2018image}, we evaluate the performance of SR solutions using PSNR and SSIM~\cite{wang2004image} on the Y channel in the transformed YCbCr space. The results are shown in Table~\ref{tab: benchmark public}, and give a reference indication on the performances of these methods on public datasets. With the same setup, we benchmark the aforementioned SR methods on our DSR test set. To quantitatively explore the effects of altitude, we evaluate separately on each altitude. The results are shown in Table~\ref{tab: benchmark a2dsr}. 
Comparing the results of the eight benchmarked SR networks on public datasets and our DSR, we see that the SR networks suffer a performance drop on DSR. Only the Urban100 shows lower PSNR and SSIM results, as this dataset contains a large number of edges and high frequency urban images. Such a drop in performance is commonly observed in real-world scenarios~\cite{wang2021unsupervised,shocher2018zero}, and the degradation model of DSR is challenging. It is also interesting to observe the variation in performance across altitudes, and that the ranking of different SR methods actually changes between datasets.

\begin{figure}[h!]
    \centering
    \includegraphics[width=\textwidth]{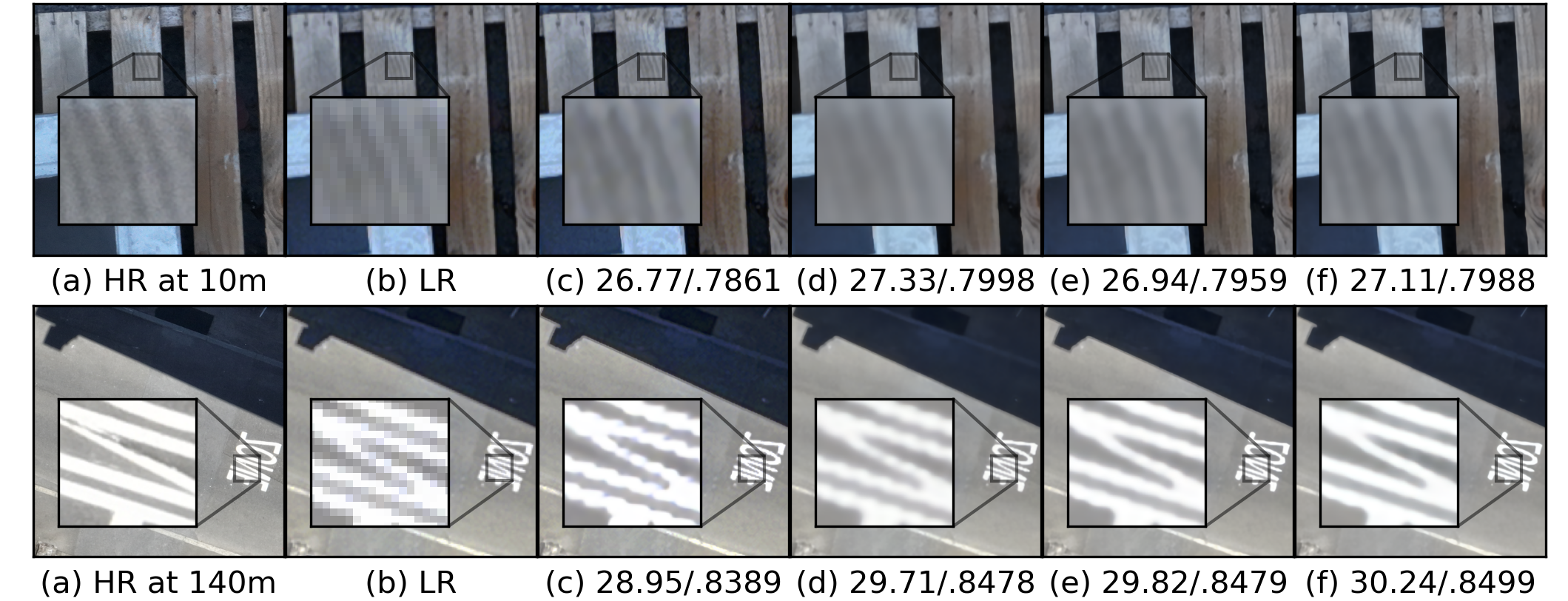}
    \caption{Qualitative and quantitative (PSNR (\textit{dB})/SSIM) results of pretrained and fine-tuned SR networks on the DSR test images. Per row: (a) HR ground-truth, (b) LR counterpart, (c) Output of SwinIR pretrained on Div2K, (d) Output of SwinIR fine-tuned on DSR 10$m$ data, (e) Output of SwinIR fine-tuned on DSR 140$m$ data, (f) Output of SwinIR fine-tuned on all DSR altitudes ($+0.34dB$ and $+1.29dB$ improvement over the baseline model).}
    \label{fig: fine-tune}
\end{figure}

\subsection{Fine-tuned SR methods}
\label{sec: fine-tune}
We further evaluate the performance of learning-based networks on DSR at various altitudes with a retraining step. We fine-tune the SwinIR~\cite{liang2021swinir} network on DSR as it achieves the best performance on the public datasets. To fit the scale factor, we add an upsampling layer followed by a convolutional layer with LeakyReLU activation at the end of the upsampling block of the original SwinIR. Following the setup in SwinIR~\cite{liang2021swinir}, We use the $\ell1$ loss with the ADAM~\cite{kingma2014adam} optimizer for fine-tuning. We first pretrain the network on Div2K synthetic image pairs, downsampled with MATLAB bicubic downsampling. Then, we fine-tune the pretrained network on all altitudes. We also repeat this configuration with a simple and fully convolutional network we refer to as FCNN~\cite{shocher2018zero}. It contains 8 hidden layers, each with 128 channels. We use ReLU activations between layers and only learn the residual. The detailed training procedure is presented in the supplementary material. 
Finally, we evaluate the performance of those fine-tuned networks as well as the pretrained network at each altitude, using PSNR and SSIM~\cite{wang2004image} on the Y channel in the transformed YCbCr space. Numerical results are shown in Table~\ref{tab: results}. Qualitative results are given in Fig.~\ref{fig: fine-tune}. We can clearly observe a significant improvement in performance with our fine-tuning on DSR, consistently over $+0.6dB$.

\begin{figure}[t]
    \centering
    \includegraphics[trim={0 0 0 0},clip,width=.95\linewidth]{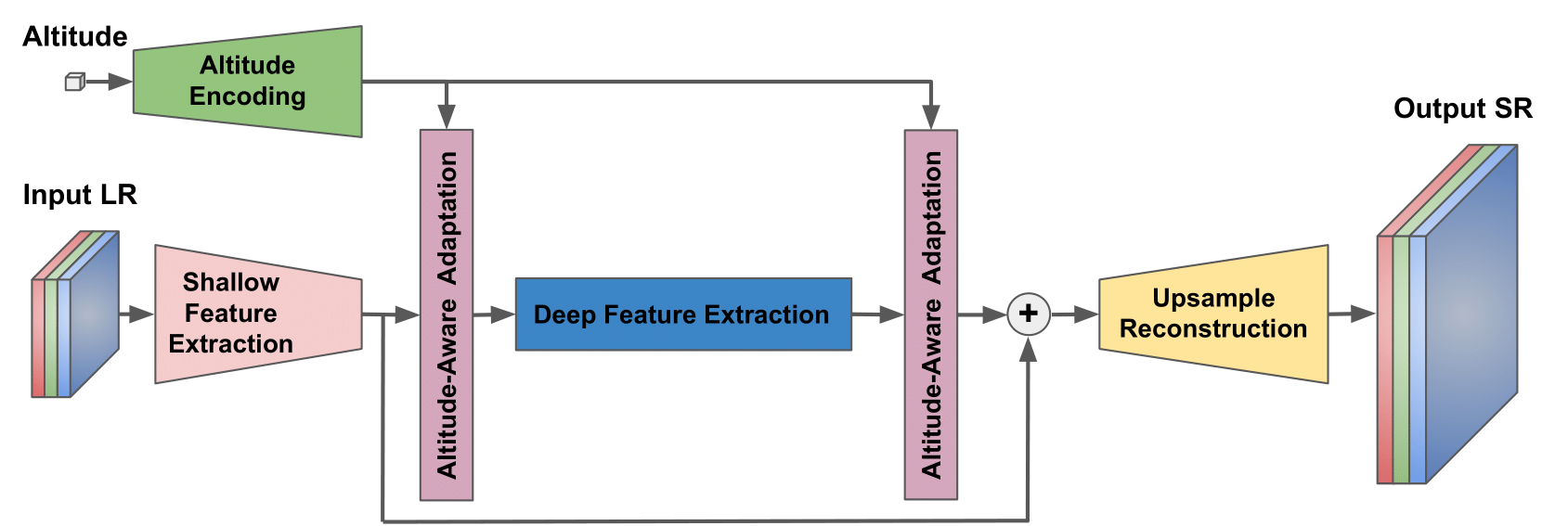}
    \caption{The architecture of our proposed altitude-aware SwinIR (AASwinIR) for drone SR. DAB from DASR~\cite{wang2021unsupervised} is used for altitude-aware feature adaptation. The altitude is first enmbedded with an encoding step, and it is then used to modify internal feature extraction. A similar modification is made over FCNN, as described in the text.}
    \label{fig: arch}
\end{figure}

\subsection{Altitude-aware SR}
The domain gaps between images captured at different altitudes (see Fig.~\ref{fig: psd}), motivates us to build a method for incorporating altitude information, which we can obtain as meta data. 
Inspired by internal feature manipulation techniques~\cite{el2020blind,lin2021fidelity} and DASR~\cite{wang2021unsupervised}, we use a degradation-aware convolutional block (DAB) for altitude awareness. The core of DAB is the degradation-aware layer (DAL), which predicts a kernel of depth-wise convolution and attention weights for channel-wise adaptation conditioned on the encoded altitude. Specifically, the encoded altitude is fed into two full-connected (FC) layers and then reshaped to a convolutional kernel. Next, the input image feature is convolved (depth-wise) with the obtained kernel. The resulting feature is sent to a 1$ \times$1 convolutional layer. The encoded altitude is also fed into the other two FC layers, followed by a sigmoid activation to generate weights for channel-wise attention. The input image feature is processed by a channel attention operation using the generated weights. The results from depth-wise convolution and channel-wise attention are summed together to obtain the output of DAL. The DAB consists of two DALs, with each DAL followed by a $3\times3$ convolutional layer.

In our altitude-aware SwinIR (AA-SwinIR), the shallow and deep image features of SwinIR~\cite{liang2021swinir}, along with the encoded altitude information, act as an input for DABs. We use two FC layers to encode the altitude. The detailed architecture of our AA-SwinIR is shown in Fig.~\ref{fig: arch}. We also modify the FCNN network to incorporate altitude information by adding one AAL between each standard convolutional layer to build our altitude-aware FCNN network (AA-FCNN). 



\input{tables/altitude-swinir}

\begin{figure}[ht]
    \centering
    \includegraphics[width=\textwidth]{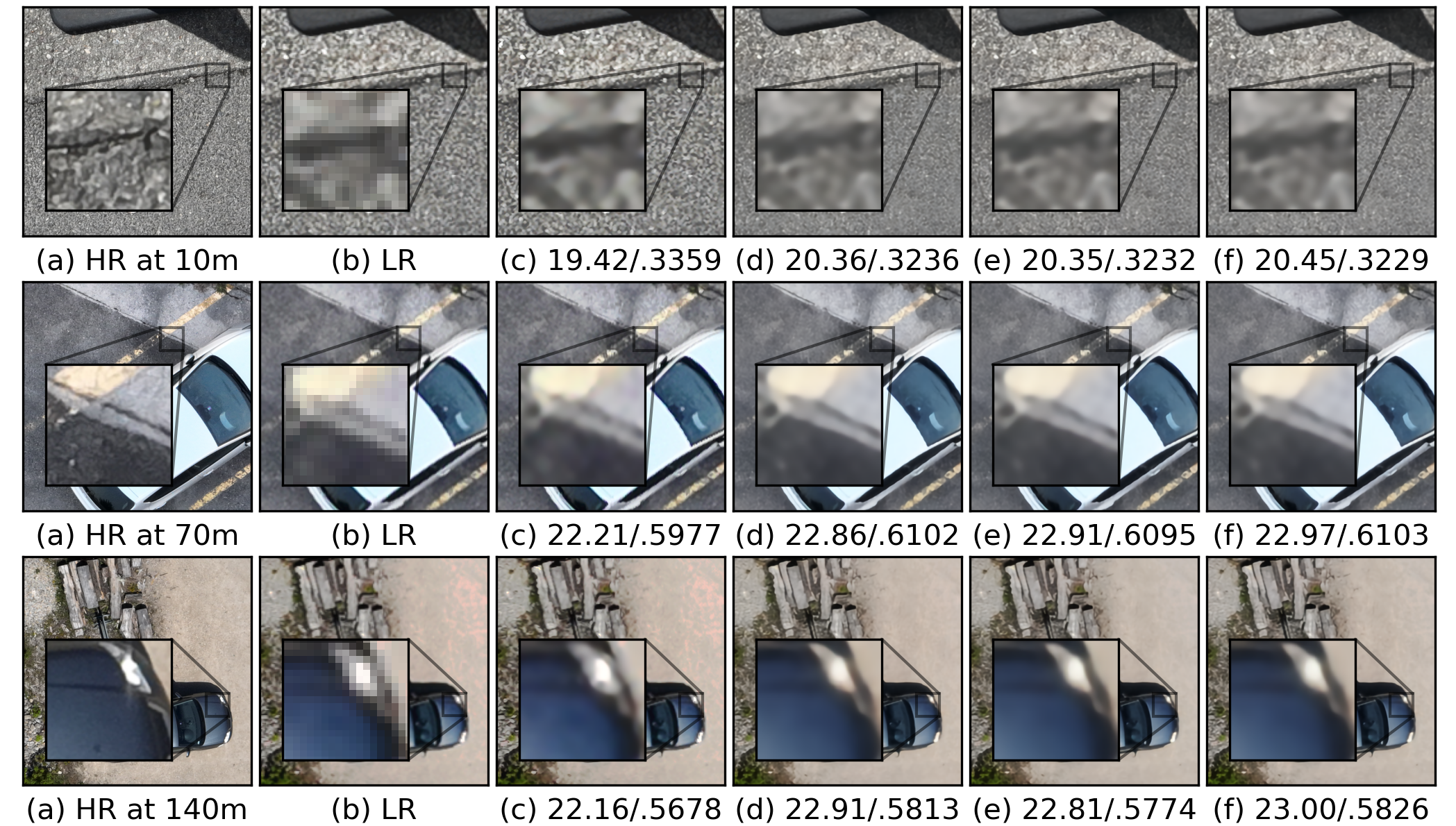}
    \caption{Qualitative and quantitative (PSNR (\textit{dB})/SSIM) results of pretrained and fine-tuned altitude-aware SwinIR network on the DSR test images. (a) HR ground-truth, (b) LR counterpart, (c) SR output of the pretrained altitude-aware AASwinIR network on Div2K, (d) SR output of SwinIR fine-tuned on all ten DSR altitudes (the same as (f) in Fig.~\ref{fig: fine-tune}), (e) SR output of AASwinIR fine-tuned on DSR without feeding altitude information, (f) SR output of the fine-tuned AASwinIR on DSR with correct altitude information.}
    \label{fig: altitude}
\end{figure}



We first pretrain our altitude-aware SwinIR network on Div2K using the same setup presented in Section~\ref{sec: fine-tune}. Since Div2K does not contain altitude information, we set all altitudes to be 1. We fine-tune this network on DSR along with the altitude information. Additionally, we normalize the altitude values by a factor of 80. The detailed training setup is presented in the supplementary material. The quantitative results of the altitude-aware AA-SwinIR pretrained on Div2K, and the fine-tuned version with altitude information are reported in Table~\ref{tab: altitude-aware}. We also report the results where the altitude is frozen at training and inference to a default value (i.e. feed in a constant 1). 

Comparing the results in Table~\ref{tab: results} and Table~\ref{tab: altitude-aware}, our fine-tuning and altitude-aware networks consistently improve the performance of SwinIR across altitudes, to a further extent with the altitude information. The improvement with the altitude-aware approach is largest at the altitudes where the fine-tuned SwinIR struggles the most (i.e. lower altitudes). However, both our fine-tuned and altitude-aware methods significantly boost the performance, up to nearly $+0.7dB$. Sample qualitative results are illustrated in columns (e) and (f) in Fig.~\ref{fig: altitude}. The quantitative results of AASwinIR \textit{without} altitude information are similar to the standard fine-tuned SwinIR, which reveals that altitude information (and not the architectural change) contributes to the improvement. We show corresponding qualitative examples in Fig.~\ref{fig: altitude} for this altitude-ablation study.




\section{Conclusion and future directions}
We create the first drone super-resolution dataset (DSR) with paired low-resolution and optical zoom high-resolution images. The paired data are captured for each scene at ten different altitudes. We show that off-the-shelf SR methods cannot be readily applied to drone images without a drop in performance, due to a domain gap in the learned priors. Furthermore, we show that a domain gap in terms of frequency content also emerges across varying altitudes. We show that fine-tuning previous methods on drone data improves the results, and even more so with our proposed altitude-aware architecture. \\ 
To foster future research towards drone super-resolution, our dataset also includes burst sequences of seven images for each of our LR captures. Future work involves extension to multi-image super-resolution, as opposed to single-image super-resolution, as well as onboard-drone efficient super-resolution solutions~\cite{bhardwaj2022collapsible,wang2022repsr,zhang2021edge}. We also carefully selected the capture altitudes such that they mutually form different multiples of each other. This can be exploited to pursue cross-altitude super-resolution methods that do not rely on pixel-level matching, but rather on learning distribution  transformation.


\clearpage
\bibliographystyle{splncs04}
\bibliography{egbib}

\end{document}

%% file: tables/statistics.tex
\newcolumntype{A}{>{\centering}p{0.2\textwidth}}
\newcolumntype{B}{>{\centering}p{0.05\textwidth}}
\newcolumntype{C}{>{\centering\arraybackslash}p{0.2\textwidth}}
\begin{table}[tb]
    \centering
    \begin{tabular}{A B B B B B B B B B B C}
        \toprule
            \multirow{2}{*}{Data split} & \multicolumn{10}{c}{\# Valid pairs at different altitudes ($m$)} & \multirow{2}{*}{Total pairs}\\
            
            & 10 & 20 & 30 & 40 & 50 & 70 & 80 & 100 & 120 & 140 &\\\hline
         
            Training & 484 & 513 & 508 & 505 & 509 & 521 & 528 & 530 & 537 & 540 & 5175 \\
            Validation & 256 & 262 & 258 & 259 & 262 & 266 & 266 & 265 & 261 & 267 & 2622 \\
            Test  & 228 & 239 & 235 & 248 & 248 & 259 & 270 & 270 & 270 & 265 & 2532 \\ \hline 
        \bottomrule
    \end{tabular}
    \caption{The final number of valid low-resolution and high-resolution image pairs in our DSR dataset for each drone capture altitude (from 10$m$ to 140$m$ with respect to the ground take-off level). The dataset is split into non-overlapping training, validation, and test sets. We note that image pairs are evenly distributed across the capture altitudes and do not create data distribution imbalance.}
    \label{tab: statistic}
\end{table}

%% file: tables/benchmark_public.tex
\newcolumntype{A}{>{\centering}p{0.2\textwidth}}
\newcolumntype{B}{>{\centering}p{0.15\textwidth}}
\newcolumntype{C}{>{\centering\arraybackslash}p{0.15\textwidth}}

\begin{table}[tb]
    \centering
    \begin{tabular}{ABBBBC}
        \toprule
        \multirow{2}{*}{Method} & \multicolumn{5}{c}{Public datasets} \\ \cline{2-6}
        & Div2K & Set5 & Set14 & B100 & Urban100 \\\hline
        
        Bicubic 
        & 26.56/.9593
        & 26.37/.7504
        & 24.30/.7065
        & 24.62/.6013
        & 21.82/.7722
        \\\hline
        
        EDSR~\cite{lim2017enhanced} 
        & 28.27/\textcolor{blue}{.9796}
        & \textcolor{red}{28.91}/.8581
        & 26.07/.7870
        & 25.80/.6657
        & 23.76/.8793
        \\
        
        RDN~\cite{zhang2018residual}
        & 28.20/.9791
        & 28.73/.8540
        & 26.06/.7841
        & 25.75/.6626
        & 23.62/.8732
        \\
        
        RCAN~\cite{zhang2018image}
        & 28.27/.9795
        & 28.78/.8575
        & \textcolor{blue}{26.11}/.7867
        & 25.81/.6668
        & 23.82/.8814
        \\
        
        ESRGAN~\cite{wang2018esrgan}
        & 26.35/.9666
        & 26.66/.7976
        & 24.53/.7186
        & 24.01/.5890
        & 22.25/.8372
        \\
        
        BSRNet~\cite{zhang2021designing}
        & 27.22/.9651
        & 27.17/.7924
        & 24.99/.7309
        & 25.16/.6285
        & 22.75/.8269
        \\
        
        SwinIR~\cite{liang2021swinir}
        & \textcolor{red}{28.45}/\textcolor{red}{.9809}
        & \textcolor{blue}{28.79}/\textcolor{red}{.8630}
        & 26.10/\textcolor{red}{.7930}
        & \textcolor{red}{25.88}/\textcolor{red}{.6727}
        & \textcolor{red}{24.09}/\textcolor{red}{.8946}
        \\
        
        NLSN~\cite{Mei_2021_CVPR}
        & \textcolor{blue}{28.31}/\textcolor{blue}{.9796}
        & \textcolor{red}{28.91}/\textcolor{blue}{.8607}
        & \textcolor{red}{26.15}/\textcolor{blue}{.7891}
        & \textcolor{blue}{25.86}/\textcolor{blue}{.6689}
        & \textcolor{blue}{24.00}/\textcolor{blue}{.8862}
        \\
        
        DASR~\cite{wang2021unsupervised}
        & 27.97/.9775
        & 28.25/.8447
        & 25.83/.7769
        & 25.64/.6590
        & 23.36/.8633
        \\
        
        \hline
        \bottomrule
    \end{tabular}
    \caption{PSNR (\textit{dB})/SSIM results of various SR methods on commonly used public datasets. We highlight the best and the second best results on each dataset in red and blue, respectively.}
    \label{tab: benchmark public}
\end{table}

%% file: tables/benchmark_our.tex
\newcolumntype{A}{>{\centering}p{0.2\textwidth}}
\newcolumntype{B}{>{\centering}p{0.15\textwidth}}
\newcolumntype{C}{>{\centering\arraybackslash}p{0.15\textwidth}}
\begin{table}[tb]
    \centering
    \begin{tabular}{ABBBBC}
        \toprule
        \multirow{2}{*}{Method} & \multicolumn{5}{c}{DSR dataset} \\ \cline{2-6}
        & 10$m$ & 20$m$ & 30$m$ & 40$m$ & 50$m$ \\\hline
        
        Bicubic 
        & \textcolor{red}{23.24}/.7156
        & \textcolor{red}{23.28}/.7112
        & \textcolor{red}{23.46}/.7118
        & \textcolor{red}{23.94}/.7271
        & \textcolor{red}{23.93}/.7341
        \\
        
        EDSR
        & 23.03/.7155
        & 23.12/.7138
        & 23.29/.7151
        & 23.73/.7289
        & 23.75/.7362
        \\
        
        RDN
        & 23.04/\textcolor{blue}{.7157}
        & 23.14/\textcolor{blue}{.7139}
        & 23.30/\textcolor{blue}{.7153}
        & 23.75/.7291
        & 23.76/.7364
        \\
        
        RCAN
        & 23.01/.7154
        & 23.12/.7137
        & 23.28/.7150
        & 23.72/.7290
        & 23.73/.7361
        \\
        
        ESRGAN
        & 22.84/.7125
        & 22.96/.7111
        & 23.09/.7123
        & 23.50/.7260
        & 23.53/.7331
        \\
        
        BSRNet
        & 22.54/.7099
        & 22.68/.7103
        & 22.89/.7126
        & 23.39/.7272
        & 23.36/.7343
        \\
        
        SwinIR$^\dagger$
        & 22.97/.7150
        & 23.08/.7133
        & 23.24/.7146
        & 23.68/.7288
        & 23.69/.7357
        \\
        
        NLSN$^\dagger$
        & 23.02/.7155
        & 23.13/.7138
        & 23.30/\textcolor{blue}{.7153}
        & 23.74/\textcolor{blue}{.7292}
        & 23.75/\textcolor{blue}{.7363}
        \\
        
        DASR$^\dagger$
        & \textcolor{blue}{23.12}/\textcolor{red}{.7166}
        & \textcolor{blue}{23.20}/\textcolor{red}{.7142}
        & \textcolor{blue}{23.37}/\textcolor{red}{.7156}
        & \textcolor{blue}{23.82}/\textcolor{red}{.7297}
        & \textcolor{blue}{23.82}/\textcolor{red}{.7369}
        \\
        
        \bottomrule
    \end{tabular}

    \begin{tabular}{ABBBBC}
        \toprule
        \multirow{2}{*}{Method} & \multicolumn{5}{c}{DSR dataset} \\ \cline{2-6}
        & 70$m$ & 80$m$ & 100$m$ & 120$m$ & 140$m$ \\\hline
        
        Bicubic
        & \textcolor{red}{24.13}/.7456
        & \textcolor{red}{24.31}/.7486
        & \textcolor{red}{24.39}/.7574
        & \textcolor{red}{24.88}/.7741
        & \textcolor{red}{24.79}/.7777
        \\
        
        EDSR
        & 23.95/.7484
        & 24.12/.7506
        & 24.22/.7603
        & 24.65/.7764
        & 24.59/.7802
        \\
        
        RDN
        & 23.96/\textcolor{blue}{.7486}
        & 24.13/\textcolor{blue}{.7508}
        & 24.24/\textcolor{blue}{.7604}
        & 24.67/\textcolor{blue}{.7766}
        & 24.61/\textcolor{blue}{.7804}
        \\
        
        RCAN
        & 23.93/.7483
        & 24.10/.7505
        & 24.21/.7602
        & 24.64/.7763
        & 24.58/.7802
        \\
        
        ESRGAN
        & 23.73/.7454
        & 23.89/.7476
        & 24.00/.7572
        & 24.39/.7731
        & 24.33/.7767
        \\
        
        BSRNet
        & 23.55/.7467
        & 23.72/.7491
        & 23.85/.7596
        & 24.31/.7761
        & 24.25/.7802
        \\
        
        SwinIR$^\dagger$
        & 23.88/.7478			
        & 24.05/.7501
        & 24.16/.7598
        & 24.58/.7759
        & 24.52/.7798
        \\
        
        NLSN$^\dagger$
        & 23.95/.7485
        & 24.12/.7507
        & 24.23/\textcolor{blue}{.7604}
        & 24.66/\textcolor{blue}{.7766}
        & 24.60/.7803
        \\
        
        DASR$^\dagger$
        & \textcolor{blue}{24.03}/\textcolor{red}{.7489}
        & \textcolor{blue}{24.20}/\textcolor{red}{.7512}
        & \textcolor{blue}{24.30}/\textcolor{red}{.7608}
        & \textcolor{blue}{24.76}/\textcolor{red}{.7771}
        & \textcolor{blue}{24.68}/\textcolor{red}{.7807}
        \\
        
        \hline
        \bottomrule
    \end{tabular}
    \caption{PSNR (\textit{dB})/SSIM results of SR methods on our DSR test set. $^\dagger$These SR networks are trained with more complex but practical degradation models considering blur, downsampling and noise. We highlight the best and the second best results at each altitude in red and blue, respectively.}
    \label{tab: benchmark a2dsr}
\end{table}

%% file: tables/results.tex
\newcolumntype{Z}{>{\centering}p{0.01\textwidth}}
\newcolumntype{A}{>{\centering}p{0.19\textwidth}}
\newcolumntype{B}{>{\centering}p{0.15\textwidth}}
\newcolumntype{C}{>{\centering\arraybackslash}p{0.15\textwidth}}
\begin{table}[]
    \centering
    \begin{tabular}{ZABBBBC}
        \toprule
        \multicolumn{2}{c}{Method} & 10$m$ & 20$m$ & 30$m$ & 40$m$ & 50$m$ \\\hline
        
        \multirow{3}{*}{\rotatebox[origin=c]{90}{FCNN}}
        & Pretrain$^\ddagger$
        & 23.11/.7174
        & 23.19/.7150
        & 23.38/.7164
        & 23.83/.7305
        & 23.83/.7378
        \\
        \cdashline{3-7}
        
        & Fine-tuned$^*$
        & 23.45/.7191
        & 23.50/.7126
        & 23.68/.7131
        & 24.14/.7292
        & 24.11/.7350
        \\
        
        & Altitude-aware
        & 23.52/.7104
        & 23.54/.7028
        & 23.70/.7051
        & 24.17/.7235
        & 24.15/.7293
        \\\hline
        
        \multirow{3}{*}{\rotatebox[origin=c]{90}{SwinIR}}
        & Pretrained$^\ddagger$
        & 23.07/.7171
        & 23.17/\textcolor{blue}{.7151}
        & 23.34/.7165
        & 23.78/\textcolor{blue}{.7305}
        & 23.79/.7375
        \\ \cdashline{3-7}
        
        & Fine-tuned$^*$
        & \textcolor{blue}{23.67}/\textcolor{red}{.7219}
        & \textcolor{blue}{23.73}/\textcolor{red}{.7162}
        & \textcolor{blue}{23.92}/\textcolor{red}{.7172}
        & \textcolor{blue}{24.36}/\textcolor{red}{.7350}
        & \textcolor{blue}{24.35}/\textcolor{red}{.7421}
        \\
        
        & Altitude-aware
        & \textcolor{red}{23.74}/\textcolor{blue}{.7204}
        & \textcolor{red}{23.76}/.7147
        & \textcolor{red}{23.95}/\textcolor{blue}{.7169}
        & \textcolor{red}{24.40}/\textcolor{red}{.7350}
        & \textcolor{red}{24.38}/\textcolor{blue}{.7420}
        \\
        
        \bottomrule
    \end{tabular}
    
    \begin{tabular}{ZABBBBC}
        \toprule
        \multicolumn{2}{c}{Method} & 70$m$ & 80$m$ & 100$m$ & 120$m$ & 140$m$ \\\hline
        
        \multirow{3}{*}{\rotatebox[origin=c]{90}{FCNN}}
        & Pretrained$^\ddagger$
        & 24.04/.7500
        & 24.22/.7525
        & 24.32/.7619
        & 24.79/.7782
        & 24.71/.7818
        \\\cdashline{3-7}
        
        & Fine-tuned$^*$
        & 24.32/.7475
        & 24.51/.7503
        & 24.59/.7587
        & 25.07/.7758
        & 25.00/.7802
        \\
        
        & Altitude-aware
        & 24.34/.7428
        & 24.52/.7465
        & 24.58/.7561
        & 25.06/.7738
        & 24.98/.7782
        \\\hline
        
        \multirow{3}{*}{\rotatebox[origin=c]{90}{SwinIR}}
        & Pretrained$^\ddagger$
        & 24.00/.7497
        & \textcolor{blue}{24.16}/.7520
        & 24.27/.7615
        & 24.71/.7778
        & 24.64/.7815
        \\\cdashline{3-7}
        
        & Fine-tuned$^*$
        & \textcolor{blue}{24.61}/\textcolor{blue}{.7570}
        & \textcolor{red}{24.78}/\textcolor{blue}{.7601}
        & \textcolor{blue}{24.88}/\textcolor{blue}{.7684}
        & \textcolor{blue}{25.36}/\textcolor{blue}{.7847}
        & \textcolor{blue}{25.28}/\textcolor{blue}{.7897}
        \\
        
        & Altitude-aware
        & \textcolor{red}{24.62}/\textcolor{red}{.7572}
        & \textcolor{red}{24.78}/\textcolor{red}{.7604}
        & \textcolor{red}{24.89}/\textcolor{red}{.7698}
        & \textcolor{red}{25.38}/\textcolor{red}{.7866}
        & \textcolor{red}{25.29}/\textcolor{red}{.7920}
        \\
        
        \bottomrule
    \end{tabular}
    \caption{PSNR (\textit{dB})/SSIM results of the pretrained and fine-tuned FCNN and SwinIR networks on our DSR test set. $^\ddagger$The network is pretrained on Div2K. $^*$The network is fine-tuned on DSR using all altitudes. We highlight the best and the second best results in red and blue, respectively.}
    \label{tab: results}
\end{table}

%% file: tables/altitude-swinir.tex
\newcolumntype{A}{>{\centering}p{0.28\textwidth}}
\newcolumntype{B}{>{\centering}p{0.134\textwidth}}
\newcolumntype{C}{>{\centering\arraybackslash}p{0.134\textwidth}}
\begin{table}[t]
    \centering
    \begin{tabular}{ABBBBC}
        \toprule
        Method & 10$m$ & 20$m$ & 30$m$ & 40$m$ & 50$m$ \\\hline
        
        AASwinIR Div2K$^\ddagger$
        & 23.06/.7168
        & 23.16/.7148
        & 23.32/.7161
        & 23.76/.7301
        & 23.77/.7371
        \\
        
        Fine-tuned SwinIR
        & \textcolor{blue}{23.67}/\textcolor{red}{.7219}
        & \textcolor{blue}{23.73}/\textcolor{blue}{.7162}
        & \textcolor{blue}{23.92}/\textcolor{red}{.7172}
        & \textcolor{blue}{24.36}/\textcolor{red}{.7350}
        & \textcolor{blue}{24.35}/\textcolor{red}{.7421}
        \\
        
        AASwinIR w/o altitude
        & 23.65/\textcolor{blue}{.7209}
        & \textcolor{blue}{23.73}/\textcolor{red}{.7164}
        & 23.91/\textcolor{blue}{.7170}
        & \textcolor{blue}{24.36}/\textcolor{blue}{.7340}
        & 24.34/.7408
        \\
        
        AASwinIR
        & \textcolor{red}{23.74}/.7204
        & \textcolor{red}{23.76}/.7147
        & \textcolor{red}{23.95}/.7169
        & \textcolor{red}{24.40}/\textcolor{red}{.7350}
        & \textcolor{red}{24.38}/\textcolor{blue}{.7420}
        \\
        
        \bottomrule
    \end{tabular}

    \begin{tabular}{ABBBBC}
        \toprule
        Method & 70$m$ & 80$m$ & 100$m$ & 120$m$ & 140$m$ \\\hline
    
        AASwinIR Div2K$^\ddagger$
        & 23.97/.7492
        & 24.14/.7515
        & 24.25/.7610
        & 24.68/.7772
        & 24.62/.7811
        \\
        
        Fine-tuned SwinIR
        & \textcolor{blue}{24.61}/\textcolor{blue}{.7570}
        & \textcolor{red}{24.78}/\textcolor{blue}{.7601}
        & \textcolor{blue}{24.88}/\textcolor{blue}{.7684}
        & \textcolor{blue}{25.36}/\textcolor{blue}{.7847}
        & \textcolor{blue}{25.28}/\textcolor{blue}{.7897}
        \\
        
        AASwinIR w/o altitude
        & 24.58/.7552
        & \textcolor{blue}{24.74}/.7581
        & 24.85/.7667
        & 25.35/.7832
        & 25.26/.7882
        \\
        
        AASwinIR
        & \textcolor{red}{24.62}/\textcolor{red}{.7572}
        & \textcolor{red}{24.78}/\textcolor{red}{.7604}
        & \textcolor{red}{24.89}/\textcolor{red}{.7698}
        & \textcolor{red}{25.38}/\textcolor{red}{.7866}
        & \textcolor{red}{25.29}/\textcolor{red}{.7920}
        \\
        \bottomrule
    \end{tabular}
    \caption{PSNR (\textit{dB})/SSIM results of the pretrained and fine-tuned altitude-aware SwinIR (AASwinIR) on our DSR test set. $^\ddagger$The network is pretrained on Div2K. The network fine-tuned with altitude information consistently yields better performance than the standard SwinIR. Without altitude information, it achieves similar results as the standard SwinIR. We highlight the best and the second best results at each altitude in red and blue, respectively.}
    \label{tab: altitude-aware}
\end{table}